# LlamaAffinity: A Predictive Antibody‑Antigen Binding Model Integrating Antibody Sequences with Llama3 Backbone Architecture

Delower Hossain[1,4], Ehsan Saghapour[2,4], Kevin Song[3,4], Jake Y. Chen[1,2,4*]

[1] Department of Computer Science, The University of Alabama at Birmingham, Birmingham, AL, US.
[2] Department of Biomedical Informatics and Data Science, The University of Alabama at Birmingham, Birmingham, AL, US.
[3] Department of Biomedical Engineering, The University of Alabama at Birmingham, Birmingham, AL, US
[4] Systems Pharmacology AI Research Center, The University of Alabama at Birmingham, Birmingham, AL, US.

*corresponding author
   jakechen@uab.edu

**Abstract-** Antibody-facilitated immune responses are central to the body's defense against pathogens, viruses, and other foreign invaders. The ability of antibodies to specifically bind and neutralize antigens is vital for maintaining immunity. Over the past few decades, bioengineering advancements have significantly accelerated therapeutic antibody development. These antibody-derived drugs have shown remarkable efficacy, particularly in treating Cancer, SARS-Cov-2, autoimmune disorders, and infectious diseases. Traditionally, experimental methods for affinity measurement have been time-consuming and expensive. With the realm of Artificial Intelligence, in silico medicine has revolutionized; recent developments in machine learning, particularly the use of large language models (LLMs) for representing antibodies, have opened up new avenues for AI-based designing and improving affinity prediction. Herein, we present an advanced antibody-antigen binding affinity prediction model (LlamaAffinity), leveraging an open-source Llama 3 backbone and antibody sequence data employed from the Observed Antibody Space (OAS) database. The proposed approach significantly improved over existing state-of-the-art (SOTA) approaches (AntiFormer, AntiBERTa, AntiBERTy) across multiple evaluation metrics. Specifically, the model achieved an accuracy of 0.9640, an F1-score of 0.9643, a precision of 0.9702, a recall of 0.9586, and an AUC-ROC of 0.9936. Moreover, this strategy unveiled higher computational efficiency, with a five-fold average cumulative training time of only 0.46 hours, significantly lower than previous studies. LlamaAffinity defines a new benchmark for antibody-antigen binding affinity prediction, achieving advanced performance in the immunotherapies and immunoinformatics field. Furthermore, it can effectively assess binding affinities following novel antibody design, accelerating the discovery and optimization of therapeutic candidates.

1  **Introduction**

Antibodies are Y-shaped proteins generated by the immune system to detect and neutralize harmful foreign substances like bacteria and viruses known as antigens. Antibodies are categorized into five primary classes: IgG, IgA, IgM, IgE, and IgD, each serving unique functions within the immune response. IgG is the most abundant and commonly used in therapeutics, while IgM is the first responder to infection. Therapeutic antibodies [1][2] have revolutionized treatment in oncology (breast, lung, and bladder cancers), autoimmune diseases, and infectious diseases. They offer high specificity with fewer side effects compared to traditional drugs. Recent trends include bispecific antibodies, antibody-drug conjugates (ADCs) [3] [4], and nanobodies. The FDA has approved at least 13 antibody drug conjugates (ADCs), including treatments for triple-negative metastatic breast cancer (MBC) and HR-positive, HER2-negative subtypes in accordance with the AXIS Pharma report. By 2025, three additional ADCs developed by AstraZeneca, Daiichi Sankyo, and AbbVie are anticipated to receive FDA approval (Biopharma PEG).

Therapeutic antibodies are top-performing biotherapeutics, with four among the top ten best-selling drugs in 2021 [5]. The global antibody drug conjugate (ADC) market was valued at $7.35 billion in 2022 and is projected to surpass $28 billion by 2028, reflecting substantial growth (Biopharma portal report). AI-driven antibody design [6][7][8][9][10] can significantly accelerate discovery and development, overcoming the time, labor, and cost limitations of traditional methods. Specifically, studies identified the following foundational antibody design models AntiBERTa [11], AntiBERTy [12], IgFold [13], AbLang [14], AbGPT [15]. AbbVie and BigHat Biosciences launched a $355 million collaboration on December 5, 2023, to leverage BigHat's AI-driven platform for next-generation oncology and neuroscience antibody therapeutics [2]. Additionally, in December 2023, AstraZeneca and AbbVie each signed deals exceeding $200 million to collaborate with Absci and BigHat Biosciences, respectively. Both partnerships aim to leverage AI-driven antibody design platforms to accelerate the development of next-generation therapeutics [2].

In novel antibody development, an antibody's effectiveness largely depends on its interaction with the target antigen, whereas binding affinity is a key indicator of this interaction's strength. Higher binding affinity generally correlates with greater therapeutic success, making it a critical focus in antibody engineering. Despite intensive research on antibody affinity systems leveraging AI, studies have shown that current approaches, such as AntiFormer [16], MVSF-AB [17], AbAgIntPre [18], AttABseq [19], and CSM-AB [20], face limitations in performance and generalizability. While Generative AI (GenAI) holds significant promise, no dedicated studies have yet explored its potential for antibody–antigen binding affinity prediction utilizing leading large language model (LLM) backbone architectures.

In this article, we propose LlamaAffinity, a novel predictive model built on the LLaMA 3 backbone [21] architecture by integrating antibody sequence data based on the Observed Antibody Space (OAS) dataset [22]. Our approach outperformed the prior state-of-the-art (SOTA) method (AntiFormer) [16] across multiple evaluation metrics. Specifically, the model achieved an accuracy of 0.9640, an F1-score of 0.9643, a precision of 0.9702, a recall of 0.9586, and an AUC-ROC of 0.9936 and an AUC-ROC of 0.9936 for the classification task. In addition, the proposed LlamaAffinity approach delivers enhanced computational efficiency relative to existing strategies.

2  **Data and Methods**

The section serves as a procedure for building the LlamaAffinity model to classify antibody binding affinity. It involves data curation and preparation, Llama backbone architecture, training, and the model performance evaluation phase.

2.1 *Dataset*

We employed the Observed Antibody Space (OAS) dataset [22], which was curated from the official AntiFormer GitHub repository (link). Since AntiFormer is the current state-of-the-art (SOTA) model in this domain, we applied the same dataset for a fair comparison. This cohort comes pre-tokenized using the ProtBERT tokenizer (BertTokenizer) from the transformer library, with a vocabulary size of 30. It includes key attributes such as input_ids, attention_mask, and token_type_ids, representing tokenized antibody sequences. Each sample is labeled with antigen-binding affinity: low affinity (label 0) and binder (label 1). The antibody sequences comprise both heavy and light chains concatenated into a single sequence. The dataset was split into five folds using StratifiedKFold cross-validation to evaluate model performance.

2.2 *Model Architecture*

Large Language Models (LLMs) have transformed and redefined the modern artificial intelligence era. The proposed model, LlamaAffinity, is developed using the LLaMA-3 backbone architecture, which is recognized as a spectacular LLM. It takes as input token IDs, padding masks, and batch size specifications. The model configuration includes four transformer layers followed by a GlobalAveragePooling layer and fully connected dense layers. The training was conducted by utilizing the adam optimizer with a learning rate 0.0001; the loss function used was Sparse Categorical Crossentropy (as shown in Equation 1). Additional key hyperparameters are summarized in Table 1.

| Parameter | Value |
| --- | --- |
| num_layers | 4 |
| num_query_heads | 12 |
| hidden_dim | 384 |
| intermediate_dim | 192 |
| vocabulary_size | 30 |
| num_key_value_heads | 12 |
| rope_max_wavelength | 100,000 |
| rope_scaling_factor | 1 |
| layer_norm_epsilon | 1e-6 |
| dropout | 0.1 |

**Equation 1** Sparse Categorical Cross Entropy

$$\text{Loss} = -[y \cdot \log(p) + (1-y) \cdot \log(1-p)] \tag{1}$$

**Equation 2** Softmax Activation

$$\text{softmax}(z_i) = \frac{e^{z_i - \max(z)}}{\sum_{j=1}^{C} e^{z_j - \max(z)}} \tag{2}$$

2.3 *Training Phase*

We conducted the simulation with five-fold cross-validation to evaluate model robustness and generalization using StratifiedKFold split. Notably, AntiFormer conducted the same cross-validation approach. The training was accomplished on Colab GPU, using over 10 epochs iterations against each fold. The total training time across all five folds was approximately 27.48 minutes, averaging about 5.5 minutes per fold, which emphasizes the model's scalability and generalizability for large-scale applications.

2.4 *Model Evaluation*

The LlamaAffinity performance was evaluated using the following metrics: accuracy, F1-score, precision, recall, and ROC AUC. Additionally, confusion matrix analysis (Figure 2) revealed strong, accurate positive rates and low misclassification predicted False positives: 3.04% only and False negatives: 4.14%.

**Equation 3** Accuracy
$$\text{Accuracy} = \frac{TP + TN}{TP + TN + FP + FN} \tag{3}$$

**Equation 4** F1 Score
$$F_1 = \frac{2 \times Precision \times Recall}{Precision + Recall} \tag{4}$$

**Equation 5** Precision
$$\text{Precision} = \frac{TP}{TP + FP} \tag{5}$$

**Equation 6** Recall
$$\text{Recall} = \frac{TP}{TP + FN} \tag{6}$$

**Equation 7** ROC AUC Score
$$\text{ROC AUC} = \int_0^1 TPR \, d(FPR) \tag{7}$$

3 **Results**

This section reveals the performance of the proposed LlamaAffinity model. The results exhibit the model's effectiveness and generalizability. Table 1 reports the outcomes of 5-fold cross-validation, while Table 2 compares the model's performance with several state-of-the-art (SOTA) approaches. Specifically, Table 1 highlights the consistent performance of LlamaAffinity across all five folds, with an average accuracy of 0.9640, F1-score of 0.9643, precision of 0.9702, recall of 0.9586, and an exceptional ROC AUC of 0.9936. Fold 3 achieved the highest accuracy (0.9725) and recall (0.9754), whereas Fold 2 recorded the highest precision (0.9847) and ROC AUC (0.9961) (Figure .1). The minimal variation in scores across folds underscores the model's robustness. In terms of efficiency, the total training time for all five folds was approximately 27.48 minutes, with each fold averaging around 5.5 minutes, demonstrating the model's scalability for large-scale applications.

**Table 1**: Cross-validation results of the LlamaAffinity model across five folds

| Fold | Accuracy | F1_score | Precision | Recall | ROC AUC | Training (Minutes) |
|---|---|---|---|---|---|---|
| 0 | 0.9550 | 0.9548 | 0.9694 | 0.9406 | 0.9886 | 5.9600 |
| 1 | 0.9525 | 0.9533 | 0.9510 | 0.9557 | 0.9913 | 4.9559 |
| 2 | 0.9675 | 0.9674 | 0.9847 | 0.9507 | 0.9961 | 5.9042 |
| 3 | 0.9725 | 0.9728 | 0.9752 | 0.9704 | 0.9969 | 5.4559 |
| 4 | 0.9725 | 0.9730 | 0.9706 | 0.9754 | 0.9951 | 5.2030 |
| Average | 0.9640 | 0.9643 | 0.9702 | 0.9586 | 0.9936 | 27.4790 |

Table 2 further displays the comparison performance of LlamaAffinity, which achieved the highest accuracy (0.9640), F1-score (0.9643), and ROC AUC (0.9936) among all evaluated models. It outperformed AntiFormer (ROC AUC: 0.9660) and AntiBERTa (ROC AUC: 0.9340) while requiring significantly less training time (0.46 hrs vs. 0.76 hrs and 2.97 hrs, respectively). In contrast, baseline models such as the 6-layer Transformer showed considerably lower accuracy (0.7865) and ROC AUC (0.7930), highlighting the strength of the Llama3 backbone in capturing intricate antibody-antigen interactions.

Additionally, the confusion matrix provides evidence of LlamaAffinity's robustness, correctly classifying 96.96% of low-affinity samples and 95.86% of binder samples. Misclassification rates remained low, with only 3.04% false positives and 4.14% false negatives, reflecting (Figure 2) a strong balance between precision and recall.

**Table 2**: Comparison of antibody affinity prediction models across performance metrics and training time.

| Model | Accuracy | F1-Score | Precision | Recall | ROC AUC | Training (hours) |
|---|---|---|---|---|---|---|
| Transformer-6 L | 0.7865 | 0.7590 | 0.8060 | 0.7990 | 0.7930 | 0.38 |
| Transformer-12 L | 0.8011 | 0.7890 | 0.8310 | 0.8180 | 0.8290 | 0.63 |
| AntiBERTy | 0.8321 | 0.8510 | 0.9110 | 0.8910 | 0.9400 | 1.46 |
| AntiBERTa | 0.8796 | 0.8570 | 0.9080 | 0.9090 | 0.9340 | 2.97 |
| AntiFormer | 0.9169 | 0.8820 | 0.9630 | 0.9250 | 0.9660 | 0.76 |
| **LlamaAffinity** | **0.9640** | **0.9643** | **0.9702** | **0.9586** | **0.9936** | **0.46** |

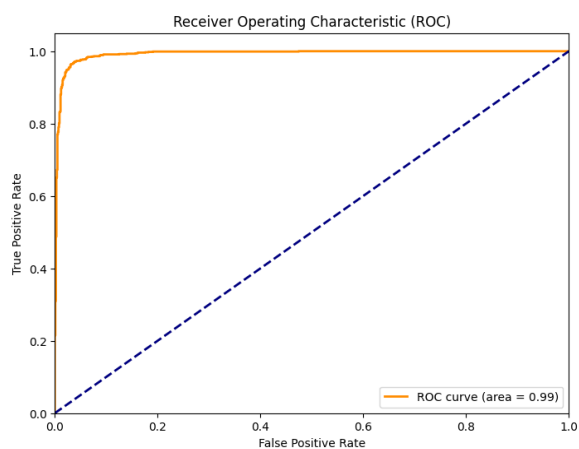 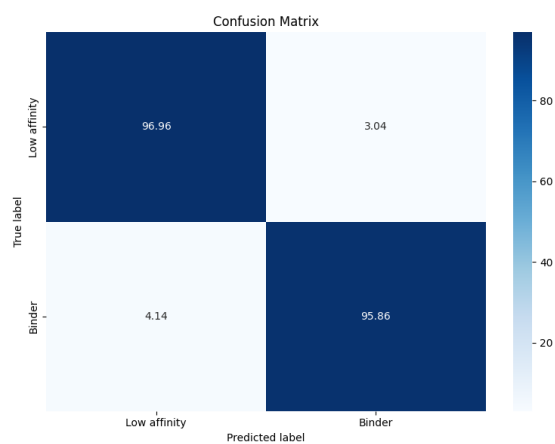

*Figure 1:* LlamaAffinity ROCAUC curve    *Figure 2:* LlamaAffinity Confusion Matrix

## 4 Conclusion

This article introduced LlamaAffinity, a novel antibody-antigen binding affinity prediction framework wrought by a Llama3 backbone architecture and antibody sequence inputs, compared to several state-of-the-art models, including AntiBERTa and AntiForme. LlamaAffinity attained the topmost performance across all evaluation metrics, with an accuracy of 0.9640, F1-score of 0.9643, and AUC-ROC of 0.9936, while maintaining a relatively low training time of 0.46 hours. These results highlight its effectiveness and computational efficiency. The proposed model advances current prediction capabilities and offers practical utility for evaluating binders in downstream novel antibody design pipelines, paving a significant step forward in immunoinformatics. For future work, conducting case studies would be ideal to validate the practical applicability of LlamaAffinity in real-world scenarios. For instance, evaluating its performance in predicting high-affinity binders for targets such as SARS-CoV-2 spike proteins or HER2 in breast cancer would provide valuable insights.

**Conflict of interest**

The authors expressed no potential conflicts of interest.

**Acknowledgements (optional)**

**Funding (optional)**

This work was partially supported by the National Institutes of Health (NIH) under grant number UMITR004771.

**Availability of data and software code (optional)**

Our code is available at the following URL: [GitHub](GitHub)

Proceedings of CIBB 2025